%% file: latex/main.tex
\pdfoutput=1

\documentclass[11pt]{article}

\usepackage[final]{acl}

\usepackage{times}
\usepackage{latexsym}
\usepackage{adjustbox}
\usepackage{makecell}
\usepackage[T1]{fontenc}

\usepackage[utf8]{inputenc}
\usepackage{graphicx}
\usepackage{subcaption}

\usepackage{microtype}
\usepackage{algorithm}
\usepackage{algorithmic}
\usepackage{amsmath}
\usepackage{inconsolata}

\usepackage{graphicx}
\usepackage{booktabs}
\usepackage{tabularx}
\usepackage{multirow}
\usepackage{multicol}
\usepackage{geometry}
\usepackage{longtable}
\title{Improving Large Language Models Function Calling and Interpretability via Guided-Structured Templates}

\author{Hy Dang\textsuperscript{1, 2\thanks{Work done while the author was an intern at Amazon.}}, Tianyi Liu\textsuperscript{2}, Zhuofeng Wu\textsuperscript{2}, Jingfeng Yang\textsuperscript{2}, Haoming Jiang\textsuperscript{2}, \\ \textbf{Tao Yang\textsuperscript{2}, Pei Chen\textsuperscript{2}, Zhengyang Wang\textsuperscript{2}, Helen Wang\textsuperscript{2}, Huasheng Li\textsuperscript{2}, }\\ \textbf{Bing Yin\textsuperscript{2}, Meng Jiang\textsuperscript{1, 2}} \\ \textsuperscript{1} University of Notre Dame, \textsuperscript{2} Amazon \\hdang@nd.edu 
}

\begin{document}

\maketitle

\begin{abstract}
Large language models (LLMs) have demonstrated strong reasoning and tool-use capabilities, yet they often fail in real-world tool-interactions due to incorrect parameterization, poor tool selection, or misinterpretation of user intent. These issues often stem from an incomplete understanding of user goals and inadequate comprehension of tool documentation. While Chain-of-Thought (CoT) prompting has proven effective for enhancing reasoning in general contexts, our analysis reveals that free-form CoT is insufficient and sometimes counterproductive for structured function-calling tasks. To address this, we introduce a curriculum-inspired framework that leverages structured reasoning templates to guide LLMs through more deliberate step-by-step instructions for generating function callings. Experimental results show that our method reduces tool-use errors, achieving 3–12\% relative improvements over strong baselines across diverse model series and approaches. Moreover, our framework enhances the robustness, interpretability, and transparency of tool-using agents, advancing the development of more reliable AI assistants for real-world applications.
\end{abstract}
\input{latex/1_intro}
\input{latex/3_method}
\input{latex/4_experiment}

\input{latex/5_results}
\input{latex/6_ablation}
\input{latex/2_related}
\input{latex/7_conclusion}
\newpage
\input{latex/9_limitation}


\label{sec:bibtex}
\bibliography{custom}
\appendix
\input{latex/8_appendix}
\end{document}

%% file: latex/1_intro.tex
\section{Introduction}

Recent advancements in large language models (LLMs), including both closed- and open-source variants, have enabled a wide range of sophisticated capabilities, including complex reasoning, planning, and tool usage \cite{touvron2023llama, liang2024survey}, targeting develops comprehensive helpful agents \cite{guo2024large, zhao2024expel, zhu-etal-2025-knowagent}. For instance, LLM-based agents can now invoke external tools or APIs to fulfill user instructions, from simple tasks (e.g., checking a date) to complex workflows (e.g., booking hotels or making purchases) \cite{qu2025tool, li-2025-review}. Despite these impressive abilities, current models frequently \textbf{fail to make correct function calls}—including errors such as incorrect parameterization, poor tool selection, or misinterpretation of user intent and hallucinations \cite{huang2023metatool, kokane2025toolscan, huang2025trustworthiness, huang2024survey}. Such failures directly impact the real-world applications, where functional correctness is critical for safety and trust \cite{xu2024reducing, zhang2024toolbehonest}.

Additionally, many LLMs \cite{qin2023toolllm, liu2024toolace, patil2024gorilla} generate function calls through a "black-box" process—providing no explanations for their selection of functions, choice of parameter values, or anticipated execution outcomes.
This \textbf{lack of explainability} significantly hinders both systematic debugging and meaningful human oversight, a critical limitation that has been widely recognized across high-stakes domains such as healthcare, finance etc.~\cite{barman2024beyond, ajwani2024llm, zhu2024explanation}. Without transparent reasoning, stakeholders cannot effectively verify the suitability of tool usage, thereby increasing the risk of consequential errors.  

Recent advances in prompting~\cite{wei2022chain, yao2023react} have shown that LLMs benefit from intermediate reasoning steps, while subsequent research~\cite{wu-etal-2024-divide, zhang2024supervised, feng2023towards, chu-etal-2024-navigate} has emphasized the importance of task-specific supervision in effectively guiding models through complex solution space. These convergent findings substantiate our central hypothesis: \textbf{LLMs require structured, contextualized guidance—not merely generic heuristics—to achieve consistent and correct tool use.} This insight represents a fundamental shift from earlier approaches that relied primarily on general-purpose prompting strategies, suggesting instead that domain-adapted scaffolding is essential for reliable tool use.

In this work, we propose a \textbf{template-based reasoning framework} for function calling that structures the model's thought process according to task demands and tool specifications. 
Our template systematically guides models through critical sub-tasks in a manner aligned with human problem-solving patterns. 
Our initial experiments revealed that while fixed structured templates substantially improve execution accuracy and instruction following, they nonetheless exhibit persistent limitations in formatting adherence, logical consistency, and functional correctness. To address these challenges, we develop a pipeline (ToolGT) to construct synthetic finetuning dataset that systematically encodes reasoning patterns using structured templates. This dataset is purposefully designed to teach models in maintaining correct formatting conventions, executing and generating step-by-step analytical reasoning, and producing outputs that precisely align with API specifications, effectively targeting the specific weaknesses observed in Template-prompting approaches.

Extensive empirical results show that our Template-based prompting and training methods on our proposed structured template consistently outperform both No-Thought and CoT approaches across models and benchmarks. On average, compared to CoT-prompting, Template-prompting achieves improvements of +2.8/+1.7 and Template-based fine-tuning yields +1.0/+1.3 on BFCLv2 and Nexus over CoT-trained models, respectively.

In summary, our contributions are two-fold:

\begin{itemize}
    \item \textbf{Template-Based Reasoning:} We develop an explicit prompting template that guide LLMs through critical stages of function calling, including tool understanding, parameter extraction, implicit conversion, and other task-specific requirements.
    \item \textbf{Structured Reasoning Dataset:} We introduce an approach for constructing the \textbf{G}uided-\textbf{T}emplate structured reasoning dataset (\textbf{ToolGT}) that effectively teaches models to improve accuracy and transparency across diverse tasks and model architectures.
\end{itemize}

We argue that equipping LLMs with curriculum-style reasoning templates offers a path toward more reliable and generalizable tool use. Rather than relying solely on unconstrained CoT reasoning, adaptive and context-specific structures can help models better align with user intent, execute accurate function calls, and provide interpretable justifications. This work establishes a foundation for future research in structured reasoning and advanced tool integration for next-generation LLM agents.

%% file: latex/3_method.tex
\section{Methodology} \label{sec:methodology}
We present a structured, template-based reasoning framework designed to enhance the function-calling capabilities of large language models (LLMs). Our approach comprises two main components: (1) prompting strategies (\S\ref{subsec:prompting_strategies}), and (2) fine-tuning strategies based on our proposed data construction method (\S\ref{subsec:sft_data_construction}).
\subsection{Problem Definition}
\label{subsection:definition}
Given a user query \( x \) and a set of tools \( T \), the function calling task traditionally aims to predict a function call \( y \) by modeling \( p(y \mid x, T) \).

We extend this by introducing a structured reasoning chain \( r \), yielding a joint modeling objective \( p(r, y \mid x, T) \), where \( r \) provides interpretable, step-by-step justification for identifying, selecting, examining, and parameterizing functions. This improves transparency and reliability, especially in function calling domains.

To generate \( r \), we aim to incorporate a guided reasoning template/curriculum \( c \) by two approaches: \textbf{prompting strategies} and \textbf{fine-tuning strategies}. In the prompting strategies, \( c \) is included in the prompt and the model will generate function call(s) based on the objective \( p(r, y \mid x, T, c) \). In contrast, supervised fine-tuning follows a data construction pipeline (\S\ref{subsec:sft_data_construction}), which generates well-curated samples that follow \( c \) during training, and at inference the model predicts \( p(r, y \mid x, T) \) without seeing \( c \).

Further details on the template/curriculum and reasoning chain construction appear in \S\ref{subsec:sft_data_construction}.

\subsection{Prompting Strategies} \label{subsec:prompting_strategies}
We first evaluate model's performance in following pre-defined set of curriculum by proposing a structured prompting methodology that guides LLMs through clearly defined reasoning steps when invoking external functions. Unlike naive CoT approaches, our method employs a structured template to enforce discrete reasoning stages.
\paragraph{Template-Based Curricula Prompting} We formalize the reasoning process into a structured curriculum or template that the model must follow before invoking any function. Inspired by how human follow specific steps to compose a function calling, the template includes: (1) Identification of functions, (2) Decision on relevancy of provided functions, (3) Examine relevant function documentation, (4) Extract and validate parameters provided by user queries, (5) Conversion of parameter types or implicit value if needed, (6) Draft a function, (7) Revalidate Function Call(s). The full detail of our template are represented in Appendix~\ref{appendix:template} (\textbf{Detail}). By embedding this structured template into the prompt, we guide the model’s reasoning process to follow specific curriculum to use tools.

\subsection{Fine-Tuning Dataset Construction} \label{subsec:sft_data_construction}

As mentioned in previous section, we internalize structured reasoning capabilities within the model. Thus, we construct a high-quality \textbf{G}uided-\textbf{T}emplate dataset (\textbf{ToolGT}) for finetuning purposes. The construction process is illustrated in \autoref{fig:data-construction-overview} and consists of:
\begin{figure*}[t!]
    \centering  \includegraphics[width=\textwidth]{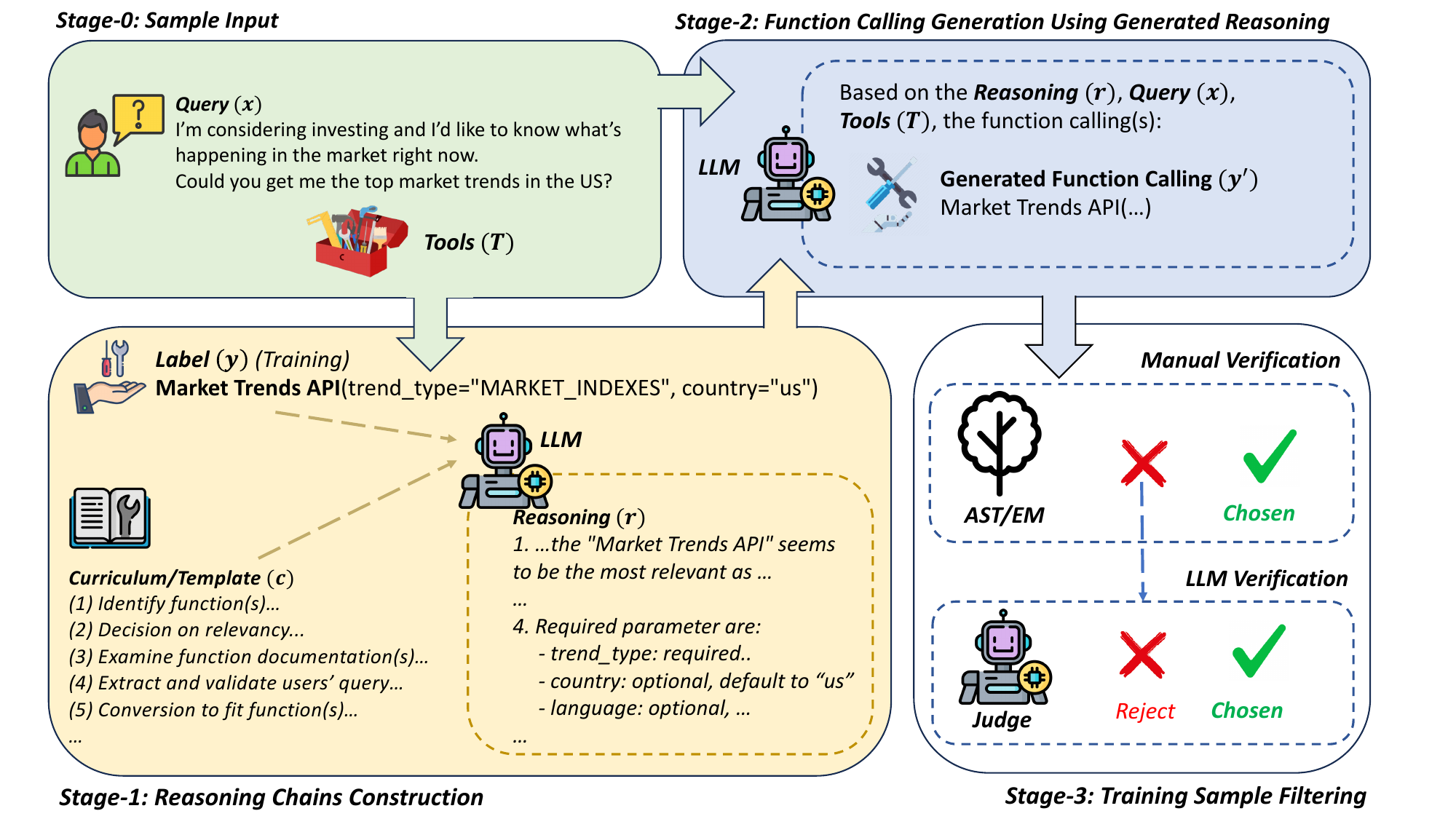}
    \caption{
    Overview of our supervised fine-tuning dataset construction \textbf{(ToolGT)} pipeline following three different stages. Stage 1 \textbf{(Reasoning Chains Construction)}—an LLM, guided by a curriculum/template $c$, generates a structured reasoning chain $r$ based on query ($x$), set of available tools $T$ and ground truth label $y$. Stage 2 \textbf{(Function Calling Generation Using Generated Reasoning)}—we evaluate the effectiveness of $r$ by prompting an LLM to predict a function call $y'$ conditioned on $(x, T, r)$, without providing $c$ at inference. Stage 3 \textbf{(Training Sample Filtering)}—to ensure high-quality supervision, we compare the predicted $y'$ with the reference $y$ using two rounds of verification: (1) Exact Match and AST-based structural comparison (AST/EM), and (2) LLM-based judgment to identify semantically equivalent alternatives. Only samples that pass verification are included in the final dataset.
    }
    \label{fig:data-construction-overview}
\end{figure*}

\paragraph{Initial Data Source} We use a tool-oriented dataset, \textbf{ToolACE} \cite{liu2024toolace}, originally developed in multi-turn dialogues. Due to their conversational nature, these datasets often include multiple unrelated turns. To improve clarity and focus, we convert them into single-turn samples, each centered on a specific function-calling instance. Specifically, we use the ToolACE subset from HuggingFace\footnote{\url{https://huggingface.co/datasets/Team-ACE/ToolACE}}, comprising 11,300 conversations.

After filtering and reformatting for BFCL-style function-calling, the dataset contains 11,488 single-turn samples. Following our data construction approach, we obtain \textbf{10,830 structured reasoning samples} for \textbf{ToolGT}. Each sample includes the components ($T$, $x$, $r$, $y$) from \S\ref{subsection:definition}, and is generated using advanced LLMs guided by our structured template. All reasoning chains are then validated to ensure high data quality.

\paragraph{Generating Structured Reasoning Chains} To generate high-quality reasoning sequences (\( r \)), we leverage advanced LLMs (e.g., GPT-4o-mini). Given the user query ($x$), the tool set ($T$), and the ground-truth function call ($y$), these models produce step-by-step reasoning chains explicitly guided by our template.

\paragraph{Function Calling Generation Using Generated Structured Reasoning} Our goal is to obtain a good reasoning chain, thus each reasoning chain $(r)$ is validated by feeding it back into the LLM alongside with $(T,x)$ to obtain $y'$ as $y' = \texttt{LLM}(x,T,r)$. And reasoning chain is kept if either method passes the training sample filtering process.
\paragraph{Training Sample Filtering} To construct a high-quality fine-tuning dataset, we compare each generated function call ($y'$), generated using constructed reasoning chain $r$ in previous step(s) to the ground truth ($y$) using a two-stage verification process:
\begin{enumerate}
\item \textbf{Manual Verification:} We apply Exact Match (EM) and Abstract Syntax Tree (AST) checks to verify syntactic and semantic equivalence. In more details, first, we compare generated with ground-truth function callings using EM approach. In the case that was marked as incorrect with EM, we figure out there are cases that the function-callings are still correct when orders of parameters are not in exact orders as in the ground truth function or there are optional parameter(s) passed but does not affect the execution. Thus, we perform additional steps of AST check to guarantee we cover these cases. 
\item \textbf{LLM-based Verification:} Given the possibility of multiple valid outputs, we use an LLM—an increasingly evaluation tool~\cite{zheng2023judging, wang2025can}—to judge if $y'$ sufficiently answers the original query ($x$), even if it differs from $y$.
\end{enumerate}
This filtering process ensures that the final fine-tuning dataset includes high-quality examples, covering valid responses—supporting robust reasoning and reliable function calling.
\subsection{Finetuning Strategies}
We train open-source models in the supervised fine-tuning (SFT) manner for our task, based on the constructed dataset described in \S\ref{subsec:sft_data_construction}. The models are initialized from publicly available pretrained checkpoints hosted on Hugging Face\footnote{\url{https://huggingface.co/}}. We experiment with several model families, as detailed in \S\ref{sec:experimental_setup}. Additional training details—including hyperparameters (see \S\ref{appendix:template}) and formatting templates (see \S\ref{appendix:data_construction})—are provided in Appendix~\ref{sec:appendix}.

%% file: latex/4_experiment.tex
\section{Experimental Setup} \label{sec:experimental_setup}

We conduct experiments to evaluate the effectiveness of explicitly embedding structured reasoning into large language models (LLMs) for enhancing their function-calling capabilities. Our evaluation focuses on two standard benchmarks: \textbf{BFCLv2\footnote{\url{https://gorilla.cs.berkeley.edu/blogs/12_bfcl_v2_live.html}}} and \textbf{Nexus \footnote{\url{https://github.com/nexusflowai/NexusRaven-V2}}}, which cover a wide range of function-calling scenarios, from simple queries to complex tasks for function calls.

It is important to note that we exclude executable test cases in the BFCLv2 benchmark from our analysis. These cases require real-time execution and external API calls, which are subject to strict rate limits. During preliminary evaluations, these constraints resulted in frequent execution errors and inconsistent results. To ensure a fair and stable comparison, we focus only on the non-executable subset of the benchmark.
\subsection{Prompting-Based Experiments}

We evaluate three prompting strategies: (1) \textbf{No Thought}, where models predict function calls without reasoning; (2) \textbf{Chain-of-Thought (CoT)}, which uses free-form reasoning ("Think-step-by-step") as guided curriculum; and (3) \textbf{Template} (ours), where models follow explicit, template-based reasoning prompts (see \S\ref{subsec:prompting_strategies}).

Our experiments include a diverse family of models, encompassing both closed- and open-source variants including:
\begin{itemize}
    \item \textbf{Closed-source models}  optimized for direct function calling include: \textbf{GPT-4o-FC} and \textbf{GPT-4o-mini-FC}, which embed templates without outputting intermediate steps. Meanwhile, \textbf{GPT-4o Prompting, GPT-4o-mini Prompting} support explicit reasoning.
    \item \textbf{Open-source} models evaluated with CoT and Template prompts/curricula come from diverse model sizes and families include: 
\textbf{LLaMA-3-70B-Instruct}, \textbf{LLaMA-3.1-8B-Instruct}, \textbf{Mistral-7B-Instruct-v0.3}, \textbf{Mistral-Nemo-12B}, and \textbf{Qwen-2.5-14B-Instruct}.
\end{itemize}

\subsection{Supervised Fine-Tuning Experiments}

To evaluate the structured reasoning on function-calling performance, we conduct supervised fine-tuning experiments using the dataset constructed following our pipeline described in \S\ref{subsec:sft_data_construction}. We compare three training strategies: 
\begin{enumerate}
    \item \textbf{No Thought} \cite{liu2024toolace}, where models were trained following ToolACE only on final function-calling;
    \item \textbf{CoT-Training}, which added free-form reasoning "Think-step-by-step" following our data construction in \S\ref{subsec:sft_data_construction};
    \item \textbf{Template-Training} (ours), which used our structured reasoning template introduce in \S\ref{subsec:prompting_strategies} with joint optimization over reasoning and function-calling.
\end{enumerate}

Experiments were run on four open-source models across different model series: \textbf{LLaMA-3.1-8B-Instruct}, \textbf{Mistral-7B-Instruct-v0.3}, \textbf{Mistral-Nemo-12B-Instruct}, and \textbf{Qwen-2.5-14B-Instruct}.

\subsection{Evaluation Metrics}
As mentioned in the previous section, we conduct our experiments using \textbf{BFCL-v2} and \textbf{Nexus}, following the metrics outlined below.
\label{subsection:evaluation_metric}
\paragraph{BFCLv2 Benchmark}:
We organize BFCL-v2 evaluation into the following sub-categories:
\begin{enumerate}
    \item \textbf{Non-Live Cases:} \texttt{simple}, \texttt{parallel}, \texttt{multiple}, \texttt{java}, \texttt{javascript}
    \item \textbf{Live Cases:} \texttt{live\_simple}, \texttt{live\_multiple}, \texttt{live\_parallel}, \texttt{live\_parallel\_multiple}
    \item \textbf{Relevancy:} \texttt{relevance}, \texttt{live\_relevance}, \texttt{live\_irrelevance}, \texttt{irrelevance}
\end{enumerate}
We compute a \emph{Weighted Average} across task categories, accounting for task frequency and complexity. For BFCL averaging, we follow the official BFCL protocol\footnote{\url{gorilla.cs.berkeley.edu/blogs/8_berkeley_function_calling_leaderboard.html}}, including \texttt{relevance} is grouped with Non-Live Categories, while \texttt{live\_relevance} and \texttt{live\_irrelevance} are grouped with Live.
\paragraph{Nexus Benchmark}: We report both a \emph{Weighted Average}, proportional to task frequency, and an \emph{Unweighted Average}, computed as a simple mean over all tasks.
These metrics capture performance across diverse task types and operational settings, reflecting both precision and robustness.

%% file: latex/5_results.tex
\section{Results and Analysis} \label{sec:results_analysis}

\begin{table}[H]
\centering
\caption{BFCL-v2 and Nexus performance using different prompting strategies for both close and open-sources models with different model families. Bold values indicate the best score per model with different strategies. Metrics are reported as Weighted / BFCL \textbf{(W / BFCL)} for BFCL-v2 and Weighted / Unweighted \textbf{(W / U)} for Nexus, following \S~\ref{subsection:evaluation_metric}.  Higher means better $\uparrow$}
\resizebox{0.5\textwidth}{!}{%
\begin{tabular}{lcc}
\toprule
\multirow{2}{*}{\textbf{Model}} & \textbf{BFCL-v2} & \textbf{Nexus} \\
  & \textbf{W / BFCL $\uparrow$} & \textbf{W / U} $\uparrow$ \\
\midrule
\multicolumn{3}{l}{\textbf{Function Calling Generation Only}} \\
\midrule
\multicolumn{3}{l}{GPT-4o-FC}\\
 + No Thought & 73.78 / 74.79 & – / – \\
 + CoT Prompting & 77.40 / 78.83 & – / – \\
 \textbf{+ Template Prompting} & \textbf{78.99} / \textbf{80.26} & – / – \\
\midrule
\multicolumn{3}{l}{GPT-4o-mini-FC}\\
 + No Thought & 73.52 / 74.72 & – / – \\
 + CoT Prompting & 77.34 / 78.71 & – / – \\
 \textbf{+ Template Prompting} & \textbf{78.30} / \textbf{79.79} & – / – \\
\midrule
\multicolumn{3}{l}{\textbf{Function Calling and Thought Generation}} \\
\midrule
\multicolumn{3}{l}{GPT-4o Prompting} \\
 + No Thought & 79.40 / 78.52 & 47.83 / 45.42 \\
 + CoT Prompting & 79.29 / 78.70 & 47.11 / 44.87 \\
 \textbf{+ Template  Prompting} & \textbf{81.21 / 80.83} & \textbf{53.18 / 51.51} \\
\midrule
\multicolumn{3}{l}{GPT-4o-mini Prompting} \\
 + No Thought & 78.99 / 80.25 & 36.85 / 34.01 \\
 + CoT Prompting & \textbf{79.87} / \textbf{80.59} & 40.61 / 35.26 \\
 \textbf{+ Template  Prompting} & 79.70 / 80.24 & \textbf{41.33} / \textbf{36.20} \\
\midrule
\multicolumn{3}{l}{LLaMA-3-70B-Instruct} \\ 
 + No Thought & 70.89 / 71.50 & 44.51 / 35.76 \\
 + CoT Prompting & 75.75 / 75.05 & 46.10 / 38.46\\
 \textbf{+ Template Prompting} & \textbf{77.78 / 76.81} & \textbf{47.40 / 41.85} \\
\midrule
\multicolumn{3}{l}{LLaMA-3.1-8B-Instruct} \\ 
 + No Thought & 64.43 / 65.75 & 35.40 / 30.06 \\
 + CoT Prompting & 65.28 / 66.70 & 36.71 / 30.06 \\
 \textbf{+ Template Prompting} & \textbf{68.28} / \textbf{69.65} & \textbf{38.01} / \textbf{33.09} \\
\midrule
\multicolumn{3}{l}{Mistral-7B-Instruct-v0.3} \\ 
 + No Thought & \textbf{60.70 / 57.55} & 10.84 / 9.05 \\
 + CoT Prompting & 51.11 / 48.94 & \textbf{26.01 / 22.26} \\
 \textbf{+ Template Prompting} & 48.83 / 46.88 & 25.29 / 20.80 \\
\midrule
\multicolumn{3}{l}{Mistral-Nemo-12B} \\ 
 + No Thought & 57.92 / 58.90 & 25.72 / 22.08 \\
 + CoT Prompting & 63.31 / 65.19 & 31.50 / 27.78 \\
 \textbf{+ Template Prompting} & \textbf{64.71} / \textbf{66.32} & \textbf{32.95} / \textbf{28.46} \\
\midrule
\multicolumn{3}{l}{Qwen-2.5-14B-Instruct} \\ 
 + No Thought & \textbf{76.22 / 77.33} & 43.21 / 38.57 \\
 + CoT Prompting & 68.06 / 62.25 & 41.33 / 35.38 \\
 \textbf{+ Template Prompting} & 74.37 / 73.28 & \textbf{44.07} / \textbf{40.78} \\
\bottomrule
\end{tabular}
}
\label{tab:bfcl_nexus_results}
\end{table}
We begin by examining the impact of Template-based reasoning through prompting strategies (see \S\ref{section:prompting_strat}), followed by fine-tuning strategies (see \S\ref{section:finetuning_strat}). 

Throughout this section, we report results using the \textit{BFCL average} for BFCLv2 and the \textit{unweighted average} for Nexus to ensure consistency and clarity. Full metric results are shown in Table~\ref{tab:bfcl_nexus_results} and Table~\ref{tab:sft_bfcl_nexus}. Additional sub-category results for both benchmarks are provided in Appendix~\ref{appendix:bfcl_eval}.

\subsection{Impact of Structured Prompting}
\label{section:prompting_strat}
\paragraph{Structured Reasoning via Template Prompting Outperforms Free-form CoT}
First, we examine \textit{whether providing a well-structured template/curriculum in the prompt can improve models' function-calling capabilities compared to no-thought or a naive CoT guidance approach}. We present the summary results in Table~\ref{tab:bfcl_nexus_results}.

Across both the BFCL-v2 and Nexus benchmarks, our results suggest that \textbf{Template Prompting} often yields better performance than CoT and No Thought, particularly across several model types and evaluation metrics on BFCL-v2 and Nexus, with further discussion on exceptions across model families provided later in this section. For example, in the function-calling generation setting, \textbf{GPT-4o-FC} achieves its best performance with Template Prompting (80.26), outperforming CoT (78.83) and No Thought (74.79). \textbf{GPT-4o-mini-FC} follows a similar trend, reaching 79.79 compared to 78.71 (CoT) and 74.72 (No Thought).

This pattern additionally holds across open-source models. Large Open-Source model \textbf{LLaMA-3-70B-Instruct} outperforms with our Template Prompting approach on both BFCLv2 and Nexus (77.78 / 47.40), outperform other counterparts, including CoT (75.75 / 46.10) and No Thought (70.89 / 35.76). 

Similarly, smaller open-source model, \textbf{LLaMA-3.1-8B-Instruct} shows notable gains with Template Prompting on both BFCL (68.28) and Nexus (33.09), outperforming CoT (66.70 / 30.06) and No Thought (65.75 / 30.06), suggesting that structured prompting more effectively guides models in tool usage. Results from \textbf{Mistral-Nemo-12B} also show benefits, with improvements of +7.42 and +1.13 on BFCL, and +6.38 / +0.68 on Nexus, compared to No Thought and CoT, respectively. These results underscore the value of structured templates in improving model's performance.

\paragraph{Template Prompting Maintains Interpretability Without Sacrificing Performance}
A noteworthy observation comes from \textbf{Qwen-2.5-14B-Instruct}, where adding reasoning steps—either through CoT or Template Prompting—lowers performance on the BFCL-v2 benchmark compared to the No Thought. We hypothesize that this is due to the model being heavily trained on direct function-calling tasks, that introducing reasoning may interfere with its learned execution patterns.

However, Template Prompting (73.28) still performs much closer to the No Thought baseline (77.33) than CoT does (62.25), highlighting the greater robustness of structured templates over free-form reasoning. Moreover, on the Nexus benchmark, Template Prompting achieves the highest score (40.78), representing relative gains of +5.7\% over No Thought (38.57) and +15.26\% over CoT (35.38). These results demonstrate that Template Prompting preserves interpretability through intermediate reasoning without compromising performance, even in high-capacity models.

\paragraph{Inconsistencies in Smaller Variants Due to Formatting and Instruction-Following Limitations.}
\label{mistral_7b}
An exception to the overall trend is \textbf{Mistral-7B-Instruct-v0.3}, which performs worse with Template-based prompting on the BFCLv2 (46.88) compared to both CoT prompting (48.94) and No Thought (57.55). Analysis of the model’s outputs suggests that this underperformance stems from its difficulty in following structured reasoning templates and formatting issues—due to limited instruction tuning or insufficient exposure to such formats during pretraining. A similar pattern is observed on the Nexus benchmark, where CoT prompting outperforms Template prompting (22.26 vs. 20.80), further emphasizing the need for targeted fine-tuning to enable smaller models to effectively leverage structured prompting strategies.

\subsection{Impact of Internalizing Structured Reasoning via Fine-Tuning}
\label{section:finetuning_strat}
As mentioned in previous section, we aim to resolve formatting and instruction-following limitations introduced by some model(s) and further enhance the model(s) performance. 

Therefore, in this section, we evaluate \textit{whether internalizing structured-reasoning models via finetuning can improve the model performance and resolve formatting and instruction-following issue(s)}. The summary results on all benchmark with experiments across different models' series is included in \autoref{tab:sft_bfcl_nexus}. 
\begin{table}[htp]
\centering
\resizebox{0.5\textwidth}{!}{%
\begin{tabular}{lcc}
\toprule
\multirow{2}{*}{\textbf{Model}} & \textbf{BFCL-v2} & \textbf{Nexus} \\
  & \textbf{W / BFCL $\uparrow$} & \textbf{W / U $\uparrow$} \\
\midrule
\multicolumn{3}{l}{LLaMA-3.1-8B-Instruct}\\
 + No Thought-Training & 69.73 / 72.16 & 35.98 / 29.82 \\
 + CoT-Training & 73.33 / 74.74 & \textbf{36.85} / 30.19 \\
 + \textbf{Template-Training} & \textbf{74.10} / \textbf{75.28} & 36.27 / \textbf{30.23} \\
\midrule
\multicolumn{3}{l}{Mistral-7B-Instruct-v0.3}\\
 + No Thought-Training & 64.65 / 66.51 & 26.59 / 21.78 \\
 + CoT-Training & 67.06 / 69.16 & 24.85 / 20.55 \\
 + \textbf{Template-Training} & \textbf{69.87} / \textbf{71.40} & \textbf{27.02} / \textbf{23.17} \\
\midrule
\multicolumn{3}{l}{Mistral-Nemo-12B-Instruct}\\
 +No Thought-Training & 74.87 / 76.44 & 32.66 / 26.46 \\
 +CoT-Training & 76.60 / 77.72 & 35.69 / 31.00 \\
 + \textbf{Template-Training} & \textbf{77.40} / \textbf{78.63} & \textbf{37.28} / \textbf{32.46} \\
\midrule
\multicolumn{3}{l}{Qwen-2.5-14B-Instruct}\\
 + No Thought-Training & 77.06 / 78.61 & 40.74 / 35.72 \\
 + CoT-Training & 77.92 / 79.46 & 40.75 / 35.85 \\
 + \textbf{Template-Training} & \textbf{78.55} / \textbf{79.83} & \textbf{42.20} / \textbf{37.05} \\
\bottomrule
\end{tabular}
}
\caption{Performance on BFCL-v2 and Nexus using fine-tuned models trained with \textbf{ToolACE}, \textbf{CoT}, or \textbf{Template-constructed (ToolGT)}. Bold values indicate the best score for each model across both benchmarks. We report the Weighted / BFCL \textbf{(W / BFCL)} average metric for BFCL-v2, and the Weighted / Unweighted \textbf{(W / BFCL)} average metric for Nexus, following the evaluation protocol described in \S~\ref{subsection:evaluation_metric}.  Higher means better $\uparrow$}
\label{tab:sft_bfcl_nexus}
\end{table}
\paragraph{Template-Based Fine-Tuning Consistently Enhances Performance and Robustness}
According to our results, template-based fine-tuning (Template-Training) consistently yields the highest performance across both benchmarks for all evaluated open-source models. 

\textbf{LLaMA-3.1-8B-Ins} improves to 75.28 with Template-Training, surpassing CoT (74.74) and the No Thought-Training baseline (72.16). \textbf{Mistral-Nemo-12B-Ins} similarly achieves its best result with Template-Training (78.63), outperforming CoT-Training (77.72) and No Thought-Training (76.44). These gains highlight the effectiveness of explicitly internalizing structured reasoning through fine-tuning. Additionally, Template-Training provides consistent guidance for function usage, improving both accuracy and interpretability in function-calling tasks compared to CoT-Training approach.

\paragraph{Training Enables Models to Effectively Utilize Structured Reasoning}
\textbf{Mistral-7B-Instruct-v0.3}, which previously underperformed with Template-prompting (\S~\ref{mistral_7b}), struggling to follow structured formats and falling below even the No Thought prompting. After Template-based fine-tuning, however, the model shows notable gains—achieving 71.40 on BFCL (vs. 66.51 for No Thought and 69.16 for CoT) and 23.17 on Nexus (vs. 21.78 for No Thought). Interestingly, fine-tuning with CoT samples results in lower Nexus performance (20.55), suggesting that unguided reasoning may hinder generalization.

These results support our earlier hypothesis: reasoning formats require dedicated training. While prompting alone may be insufficient—especially for smaller models—supervised fine-tuning enables systematic reasoning, effectively bridging the gap between structure and performance.

\paragraph{Broad Applicability Across Different Models}
Template-based training proves effective across model series. For example, \textbf{Qwen-2.5-14B-Ins} achieves top performance across both benchmarks (BFCL: 79.83; Nexus: 37.05). Notably, CoT fine-tuning yields only marginal gains over baseline in the Nexus benchmark (e.g., 35.85 vs. 35.72, respectively), whereas Template-training provides meaningful improvements (37.05), account for approximately +3.5\% relative improvements over both No Thought-Training and CoT-Training. 

These results suggest that template supervision generalizes well across models, offering a robust and interpretable foundation for enhancing tool use, particularly in complex function-calling tasks.

%% file: latex/6_ablation.tex
\section{Ablation Study}
\subsection{Impact of Template Complexity}
\label{subsec:template_ablation}

To study the effectiveness of our proposed template \textbf{Detail}. We experiment with different types of templates (\textbf{Simple, Claude, Detail}) (See \S\ref{appendix:template} for details) with results summarized in \autoref{table:bfcl_template_ablation}. 

According to \autoref{table:bfcl_template_ablation}, the \textbf{Detail} template consistently achieves the highest overall accuracy. For example, it outperforms other variants across most tasks (LLaMA-3.1-8B-Instruct: 74.10 vs. 71.27 for \textbf{Simple}, 70.20 for \textbf{Claude} in overall accuracy). Additionally, we make interesting observation related to the benefit(s) of different type of template. In more details, \textbf{Simple} occasionally performs better on specific subtasks, such as \textit{Relevancy}, suggesting that task-specific template complexity may be beneficial. 

These results highlight a trade-off between \textbf{specificity} and \textbf{simplicity}, motivating future research on \textbf{adaptive strategy} that dynamically adjust reasoning depth based on task characteristics, which we leave it for future work.

\begin{table}[htbp]
    \centering
    \caption{Performance across different templates used for training data generation with Mistral-v0.3 and Llama-3.1 models. "No Thought" denotes training directly on function calls, whereas "With Thought" explicitly includes structured reasoning. Evaluations are based on BFCL benchmark using \textit{Weighted average}. Higher means better $\uparrow$}
    \label{table:bfcl_template_ablation}
    \resizebox{\linewidth}{!}{
    \begin{tabular}{lccc}
        \toprule
        \textbf{Category} & \textbf{Simple} & \textbf{Claude} & \textbf{Detail} \\
        \midrule
        \multicolumn{4}{c}{\textbf{Mistral-v0.3-7B-Ins}} \\
        \midrule
        \multicolumn{4}{l}{\textbf{No Thought Training}} \\
         Overall Score & 66.44 & 47.73 & 64.65 \\
        \midrule
         \multicolumn{4}{l}{\textbf{With Thought Training}} \\
         Overall Score & 68.63 & 60.64 & \textbf{69.87} \\
         Non-Live Average & 78.52 & 70.26 & \textbf{81.22} \\
         Live Average & 55.35 & 47.19 & \textbf{59.92} \\
         Relevancy & \textbf{74.13} & 66.59 & 70.07 \\
        \midrule
        \multicolumn{4}{c}{\textbf{Llama-3.1-8B-Ins}}\\
        \midrule
         \multicolumn{4}{l}{\textbf{No Thought Training}} \\
         Overall Score & 64.49 & 61.55 & 62.76 \\
        \midrule
         \multicolumn{4}{l}{\textbf{With Thought Training}} \\
         Overall Score & 71.27 & 70.20 & \textbf{74.10} \\
         Non-Live Average & 80.43 & 79.57 & \textbf{84.61} \\
         Live Average & 55.58 & 54.60 & \textbf{63.68} \\
         Relevancy & \textbf{80.28} & 78.90 & 75.61 \\
        \bottomrule
    \end{tabular}}
\end{table}

\subsection{Lack of Training Samples Can Harm in Complex Scenarios.}
\label{appendix:toolace}

\begin{figure*}[htbp]
    \centering
    \includegraphics[width=\linewidth]{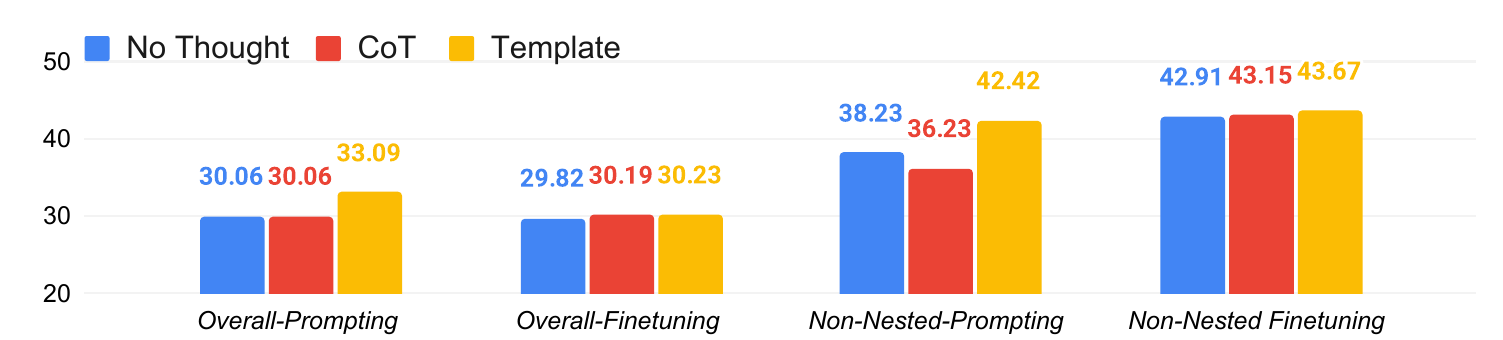}
    \caption{Performance comparison of LLaMA-3.1-8B-Instruct across prompting strategies (No Thought, CoT, Template) before and after fine-tuning (SFT) on Nexus benchmark. The results highlight performance degradation in overall scenarios (Left 2 groups), contrasted by clear improvements on the non-nested subset (Right two groups)}
    \label{fig:llama_ablation}
\end{figure*}
\begin{figure*}[htbp]
    \centering
    \includegraphics[width=\linewidth]{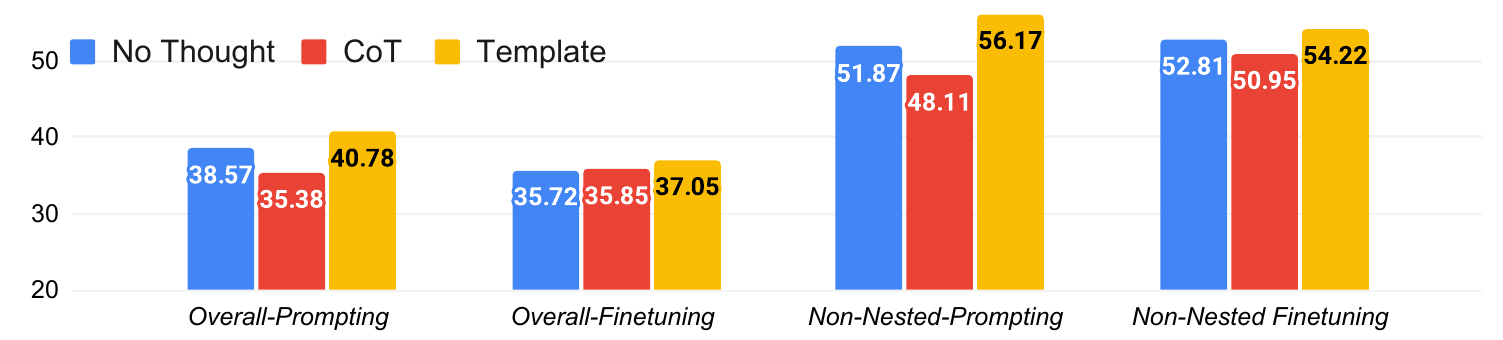}
    \caption{Performance comparison of Qwen-2.5-14B-Instruct across prompting strategies (No Thought, CoT, Template) before and after fine-tuning (SFT) on Nexus benchmark. The results highlight performance degradation in overall scenarios (Left 2 groups), contrasted by clear improvements on the non-nested subset (Right two groups)}
    \label{fig:qwen_ablation}
\end{figure*}

During our analysis, we observed unexpected performance degradation in \textbf{LLaMA-3.1-8B-Instruct} and \textbf{Qwen-2.5-14B-Instruct} after fine-tuning on ToolACE (No Thought), CoT-Training (CoT), and ToolGT (Template), across all reasoning strategies (see~\autoref{tab:bfcl_nexus_results} and \autoref{tab:sft_bfcl_nexus}) in \textbf{Nexus}. For example, in LLaMA, No Thought dropped slightly from 30.06 to 29.82, while Template prompting decreased more noticeably from 33.09 to 30.23. The drop was more pronounced in Qwen, with No Thought falling from 38.57 to 35.72—a relative decrease of about 7\%. We hypothesize that this degradation is due to ToolACE’s limited coverage of complex, nested function-call scenarios, a limitation also present in the BFCLv2 evaluation.

To investigate, we analyzed performance on a \textit{non-nested} subset of Nexus (Figure~\ref{fig:llama_ablation}, Figure~\ref{fig:qwen_ablation}). In this setting, fine-tuned models improved consistently across all prompting strategies. For instance, LLaMA’s Template-trained variant achieved the highest score (43.67), and Qwen showed gains in the first two prompting conditions. However, Qwen’s Template-trained model, while outperforming other training strategies, still lagged behind zero-shot template prompting (54.22 vs. 56.17).

These results suggest that without adequate training coverage of compositional tool use, fine-tuned models may overfit to simpler patterns, impairing performance on more complex tasks. We leave the development of datasets targeting nested and compositional reasoning for future work.

%% file: latex/2_related.tex
\section{Related Work}
\label{sec:related_work}

Research on methods that enable tool usage in language models broadly falls under two main categories: prompting-based (\S\ref{related_tool_use}) and tuning-based (\S\ref{related_tuning_tool_use}) approaches, with some recent attention to extended reasoning techniques (\S\ref{related_rl}).

\subsection{Prompting-Based Methods}
\label{related_tool_use}
Prompting methods equip LLMs with textual tool descriptions to help them select and invoke tools for tasks such as question answering~\cite{yao2023react,lu2023chameleon} and mathematical problem solving~\cite{imani-etal-2023-mathprompter, das-etal-2024-mathsensei}. More advanced prompting strategies introduce step-by-step reasoning or iterative interactions~\cite{paranjape2023art,wei2022chain,sun2023adaplanner}. However, these approaches often rely on simplistic tool definitions and require multiple generation rounds, limiting efficiency and scalability. In contrast, our work focuses on enabling accurate tool use through structured, template-guided prompting while maintaining efficiency.

\subsection{Tuning-Based Methods}
\label{related_tuning_tool_use}
Recent work has explored fine-tuning models for function calling~\cite{gao2024confucius,schick2023toolformer, qin2024tool}. For example, Toolformer~\cite{schick2023toolformer} uses specialized tokens to guide tool use, while ToolkenGPT~\cite{hao2023toolkengpt} encodes tools as tokens to generalize to unseen tools. A similar work, ToolGen~\cite{wang2410toolgen}, also aims to embed tools as tokens and tries to unify retrieval and tool-call generation. Other methods train on curated synthetic datasets from different approaches~\cite{guo2024stabletoolbench,liu2024toolace,li2023api}; for example, ToolACE~\cite{liu2024toolace} uses a multi-agent framework, while Gorilla builds on APIBench~\cite{patil2024gorilla} to support diverse APIs. While these approaches improve accuracy, they primarily map queries directly to function calls without intermediate reasoning. TRICE~\cite{qiao2023making} introduces post-execution feedback; in contrast, our approach emphasizes pre-execution structured reasoning to enhance interpretability and performance. Additionally, there is research that aims to provide some form(s) of template reasoning, such as Synthesize Step-by-Step~\cite{li2024synthesize}, which employs templates to train an LLM-based data generator that produces questions and step-by-step rationales for chart visual question answering problems. In contrast, our work targets function calling and tool-use reasoning, aiming to enhance execution correctness and interpretability during inference. 

\subsection{Extended CoT for Long Reasoning}
\label{related_rl}
Recent models such as Deepseek-R1 \cite{guo2025deepseek} adopt extended CoT reasoning to improve interpretability and robustness. However, these approaches often introduce significant token overhead, length constraints, and formatting issues, limiting their practicality in latency-sensitive applications, especially in small models \cite{feng2025efficient, guo2025deepseek, dang2025reinforcement}. Additionally, this reinforcement learning techniques requires intensive training architectures \cite{parmar2025challenges}. Thus, we propose a complementary, template-driven strategy dataset that could be used to train models to perform structured, grounded reasoning. This method maintains interpretability while reducing error rates and improving efficiency.

%% file: latex/7_conclusion.tex
\section{Conclusion} \label{sec:conclusion}
In this paper, we propose a structured, template-based approach to enhance the function-calling capabilities of LLMs. By following carefully designed templates tailored for function call generation, our method guides models through deliberate, step-by-step reasoning rather than relying on naive, unguided outputs. Experimental results demonstrate that both structured prompting and supervised fine-tuning significantly improve function call accuracy while also enhancing interpretability across various models and benchmarks. Our contributions include explicit reasoning templates for function calling, a synthetic dataset construction method, and empirical evidence of performance gains. Future directions include extending structured reasoning to broader decision-making tasks and adaptive curricula to further improve the reliability and transparency of LLM-based agents.

%% file: latex/9_limitation.tex
\section*{Limitations}
Our study has several limitations that present opportunities for future research. First, we only consider a subset of ToolACE, as only a limited portion of the dataset has been publicly released. This restricts the comprehensiveness of our evaluations and limits the potential of ToolACE in data construction. In future work, we plan to augment these cases with additional function-calling datasets—including more complex, nested scenarios that ToolACE lacks—to construct a more rigorous and diverse training corpus under our proposed framework. Second, our analysis reveals that different prompting templates can offer advantages in specific function-calling scenarios. This suggests a promising direction for developing models that can adaptively select and apply the most suitable template based on contexts and scenarios. Finally, our current experiments are limited to single-turn function-calling tasks. Currently, we do not evaluate or develop our framework to multi-turn scenarios, which introduced in BFCLv3 \footnote{\url{https://gorilla.cs.berkeley.edu/blogs/13_bfcl_v3_multi_turn.html}}. Thus, future work should incorporate multi-turn tool, function calling settings.
\section*{Acknowledgements}
This work was supported by an internship at Amazon and NSF IIS-2119531, IIS-2137396, IIS-2142827, and IIS-2234058. We also appreciate the support from the Foundation Models and Applications Lab of Lucy Institute and ND-IBM Tech Ethics Lab.

%% file: latex/8_appendix.tex
\section{Appendix}
\label{sec:appendix}

\subsection{Training \& Hyperparameters} \autoref{table:training_hyperparameters} provides the detailed hyperparameters used for supervised fine-tuning experiments. Training is conducted on  8 NVIDIA A100 80GB GPUs, requiring approximately 15 GPU
hours for each finetuning experiment.

\begin{table}[htbp]
    \centering
    \caption{Training hyperparameters for supervised fine-tuning experiments.}
    \label{table:training_hyperparameters}
\resizebox{\linewidth}{!}{
    \begin{tabular}{lc}
        \toprule
        \textbf{Hyperparameter} & \textbf{Value} \\
        \midrule
        Batch Size (per device) & 2 \\
        Thought Generation Model & GPT-4o-mini \\
        Training Sample Verification & GPT-4o-mini \\
        Learning Rate & 2e-5 \\
        LR Scheduler & cosine \\
        Warmup Ratio & 0.1 \\
        Epochs & 3 \\
        Max Sequence Length & 8192 \\
        \bottomrule
    \end{tabular}}
\end{table}

\subsection{Templates Used For Data Construction}
\autoref{table:appendix_templates} summarizes the templates (Claude, Simple, Detail) used in our training data construction and experiments, along with the number of helpful training samples generated using GPT-4o-mini. Among these, the \textbf{Detail} template serves as the primary design for both our prompting and data construction strategies. 

As discussed in \S\ref{subsec:template_ablation}, we observe varying effectiveness across templates, with the \textbf{Detail} template consistently outperforming the others. For prompting strategies, we also adopt \textbf{Detail} in our Template-Prompting experiments.
\label{appendix:template}

\begin{table}[htbp]
\centering
\label{table:appendix_templates}
\resizebox{\linewidth}{!}{
\begin{tabular}{p{1.4cm}|p{7.5cm}}
\toprule
\textbf{Template} & \textbf{Prompt} \\
\midrule
\makecell[l]{\textbf{Detail} \\ \textbf{(10,830)}} & 
1.Identify which function or set of functions best fit the user query based on function descriptions.

2.Pick that function or set of functions to fulfill the user's request.

3. After selecting the function(s), carefully examine the function documentation.

4. Analyze the provided parameters, considering descriptions, parameter types, and optionality.

5.Extract relevant information from user queries, performing necessary type conversions.

6. Draft the function call(s) fulfilling the user's request.

7. Revalidate the composed functions, ensuring they satisfy both documentation and the user's query.
\\
\midrule
\makecell[l]{\textbf{Claude\footnotemark} \\ \textbf{(9,307)}} &First, determine the most relevant tool provided to answer the user's request. Then, review each required parameter for the selected tool, verifying if the user explicitly provided or implicitly offered sufficient information for inference. If all required parameters can be directly or reasonably inferred, proceed to invoke the function. If a required parameter is missing, do not invoke the function and instead request the missing parameters from the user. Avoid requesting optional parameters if not explicitly provided.
\\
\midrule
\makecell[l]{\textbf{Simple} \\ \textbf{(10,300)}} & 
First, identify the appropriate tool(s) for answering the user's request. Then analyze the query to formulate the necessary function call parameters. If no suitable tools are available or insufficient information is provided, refrain from making a function call.
\\
\bottomrule
\end{tabular}}
\caption{Template-wise statistics of helpful training samples used during dataset creation with structured prompts.}
\label{table:appendix_templates}
\end{table}
\footnotetext{\url{https://docs.anthropic.com/en/docs/agents-and-tools/tool-use/overview\#chain-of-thought-tool-use}}

\subsection{Detailed Experiment Results}

\label{appendix:bfcl_eval}
\paragraph{BFCL-v2 Evaluation Details} For this section, we present the full BFCLv2 results of our experiments including both prompting and finetuning strategies in \autoref{tab:bfcl_breakdown}. 
\begin{table*}[htbp]
\centering
\caption{Sub-category results on BFCL-v2 using prompting and fine-tuning strategies. Bold values indicate the best performance per model. Averages of each categories computed using the weighted metric. Meanwhile, \textbf{W / BFCL} average metric still follows evaluation protocol mentioned in  \S\ref{subsection:evaluation_metric}}
\begin{adjustbox}{width=0.8\textwidth}
\begin{tabular}{lcccc}
\toprule
\textbf{Model} & \textbf{Non-Live Cases} & \textbf{Live-Cases} & \textbf{Relevancy} & \textbf{W / BFCL $\uparrow$} \\
        \midrule
        \multicolumn{5}{c}{\textbf{Prompting Strategies}} \\
        \midrule
\textbf{GPT-4o-FC} &  &  &  &  \\
+No Thought & 84.87 & 69.81 & 67.32 & 73.78 / 74.79 \\
+CoT & 86.00 & 63.45 & 84.95 & 77.40 / 78.83 \\
+Template & 87.65 & 66.51 & 84.78 & \textbf{78.99 / 80.26} \\
\midrule
\textbf{GPT-4o-mini-FC} &  &  &  &  \\
+No Thought & 85.22 & 69.36 & 66.70 & 73.52 / 74.72 \\
+CoT & 84.52 & 64.57 & 84.95 & 77.34 / 78.71 \\
+Template & 86.78 & 67.57 & 82.27 & \textbf{78.30 / 79.79} \\
\midrule
\textbf{GPT-4o-mini-Prompting} &  &  &  &  \\
+No Thought & 88.78 & 73.71 & 75.34 & 78.99 / 80.25 \\
+CoT & 85.91 & 72.58 & 82.26 & \textbf{79.87 / 80.59} \\
+Template & 86.43 & 71.99 & 81.92 & 79.70 / 80.24 \\
\midrule
\textbf{LLaMA-3.1-8B-Instruct} &  &  &  &  \\
+No Thought & 86.35 & 66.37 & 40.40 & 64.43 / 65.75 \\
+CoT & 81.91 & 62.25 & 52.25 & 65.28 / 66.70 \\
+Template & 84.96 & 65.10 & 55.36 & \textbf{68.28 / 69.65} \\
\midrule
\textbf{Mistral-7B-Instruct-v0.3} &  &  &  &  \\
+No Thought & 64.17 & 54.60 & 64.27 & \textbf{60.70 / 57.55} \\
+CoT & 56.61 & 54.75 & 41.43 & 51.11 / 48.94 \\
+Template & 57.57 & 51.01 & 37.63 & 48.83 / 46.88 \\
\midrule
\textbf{Mistral-Nemo-12B-Instruct} &  &  &  &  \\
+No Thought & 86.17 & 72.96 & 12.46 & 57.92 / 58.90 \\
+CoT & 84.09 & 66.59 & 38.84 & 63.31 / 65.19 \\
+Template & 83.91 & 68.09 & 41.70 &\textbf{ 64.71 / 66.32} \\
\midrule
\textbf{Qwen-2.5-14B-Instruct} &  &  &  &  \\
+No Thought & 88.96 & 70.04 & 70.68 & \textbf{76.22 / 77.33} \\
+CoT & 61.57 & 70.26 & 71.97 & 68.06 / 62.25 \\
+Template & 79.22 & 71.38 & 73.01 & 74.37 / 73.28 \\
\midrule
\multicolumn{5}{c}{\textbf{Finetuning Strategies}} \\
\midrule
\textbf{LLaMA-3.1-8B-Instruct} &  &  &  &  \\
+No Thought & 87.22 & 63.22 & 59.86 & 69.73 / 72.16 \\
+CoT & 82.70 & 63.14 & 75.78 & 73.33 / 74.74 \\
+Template & 84.61 & 63.68 & 75.61 & \textbf{74.10 / 75.28} \\
\midrule
\textbf{Mistral-7B-Instruct-v0.3} &  &  &  &  \\
+No Thought & 83.39 & 64.27 & 46.45 & 64.65 / 66.51 \\
+CoT & 79.39 & 59.02 & 64.10 & 67.06 / 69.16 \\
+Template & 81.22 & 59.92 & 70.07 & \textbf{69.87 / 71.40} \\
\midrule
\textbf{Mistral-Nemo-12B-Instruct} &  &  &  &  \\
+No Thought & 85.04 & 63.82 & 77.51 & 74.87 / 76.44 \\
+CoT & 84.52 & 68.92 & 77.60 & 76.60 / 77.72 \\
+Template & 86.78 & 68.54 & 78.31 & \textbf{77.40 / 78.63} \\
\midrule
\textbf{Qwen-2.5-14B-Instruct} &  &  &  &  \\
+No Thought & 88.09 & 67.19 & 77.51 & 77.06 / 78.61 \\
+CoT & 87.04 & 68.83 & 79.32 & 77.92 / 79.46 \\
+Template & 88.70 & 69.22 & 79.24 & \textbf{78.55 / 79.83} \\
\bottomrule
\end{tabular}
\end{adjustbox}
\label{tab:bfcl_breakdown}
\end{table*}
\paragraph{Nexus Evaluation Details} 

\begin{table*}[htbp]
    \centering
    \caption{Detailed results on sub-categories of the Nexus Function Calling benchmark using prompting and fine-tuned strategies on open-source models. \textbf{VT} refers to the VirusTotal API, while \textbf{VT (N)} and \textbf{VT (P)} represent nested and parallel cases, respectively. Bold values indicate the best for each model within each setting. Averages are computed using the weighted/unweighted \textbf{(W/U) mean} in \S\ref{subsection:evaluation_metric}.}
    \begin{adjustbox}{width=\textwidth}
    \begin{tabular}{lcccccccccc}
        \toprule
        \textbf{Model} & \textbf{NVDLibrary} & \textbf{VT} & \textbf{Places} & \textbf{Climate} & \textbf{OTX} & \textbf{VT (N)} & \textbf{VT (P)} & \textbf{CVECPE} & \textbf{W/U $\uparrow$} \\
        \midrule
        \multicolumn{9}{c}{\textbf{Prompting Strategies}} \\
        \midrule
        \multicolumn{9}{l}{\textbf{LLaMA-3.1-8B-Instruct}}\\
        +No Thought                   & 38.46 & 68.87 & 16.67 & 9.64 & 82.61 & 8.16 & 14.29 & 1.79 & 35.40 / 30.06 \\
        +CoT                    & 50.00 & 66.89 & 8.33  & 12.18 & 84.78 & 8.16 & 4.76  & 5.36 & 36.71 / 30.06 \\
        +Template               & 43.59 & 68.87 & 18.75 & 13.20 & 83.70 & 12.24 & 19.05 & 5.36 & \textbf{38.01 / 33.09} \\
        \midrule
        \multicolumn{9}{l}{\textbf{Mistral-7B-Instruct-v0.3}}\\
        +No Thought              & 24.35 & 12.58 & 4.17  & 7.11 & 20.65 & 0.00 & 0.00 & 3.57 & 10.84 / 9.05 \\
        +CoT                    & 35.90 & 48.34 & 8.33  & 5.58 & 78.26 & 0.00 & 0.00 & 1.79 & \textbf{26.01 / 22.26} \\
        +Template               & 41.03 & 44.37 & 8.33  & 6.09 & 63.04 & 0.00 & 0.00 & 3.57 & 25.29 / 20.80\\
        \midrule
         \multicolumn{9}{l}{\textbf{Mistral-Nemo-12B-Instruct}}\\
         
        +No Thought            & 41.03 & 37.09 & 8.33  & 6.09 & 79.35 & 0.00 & 4.76 & 0.00  & 25.72 / 22.08\\
        +CoT                    & 42.31 & 56.29 & 10.42 & 5.58 & 84.78 & 2.04 & 19.05 & 1.79 & 31.50 / 27.78\\
        +Template               & 50.00 & 58.94 & 14.58 & 7.61 & 80.43 & 0.00 & 14.29 & 1.79 & \textbf{32.95 / 28.46}\\
        \midrule
         \multicolumn{9}{l}{\textbf{Qwen-2.5-14B-Instruct}}\\
         
        +No Thought            & 60.25 & 79.47 & 29.17 & 12.69 & 90.22 & 8.16 & 28.57 & 0.00 & 43.21 / 38.57 \\
        +CoT                    & 73.07 & 78.14 & 10.42 & 9.14  & 89.13 & 4.08 & 19.05 & 0.00 & 41.33 / 35.38 \\
        +Template               & 79.49 & 76.16 & 25.00 & 10.15 & 89.13 & 4.08 & 33.33 & 8.93 & \textbf{44.07 / 40.78}\\
        \midrule
        \multicolumn{9}{c}{\textbf{Finetuning Strategies}} \\     
        \midrule
        \multicolumn{9}{l}{\textbf{LLaMA-3.1-8B-Instruct}}\\
        +No Thought                   & 46.15 & 69.54 & 36.36 & 85.00 & 88.04 & 0.00 & 18.18 & 0.00 & 35.98 / 29.82 \\
        +CoT                    & 62.82 & 70.86 & 45.45 & 90.00 & 76.09 & 0.00 & 0.00  & 0.00 & \textbf{36.85 }/ 30.19 \\
        +Template               & 66.67 & 60.93 & 40.91 & 95.00 & 85.87 & 0.00 & 0.00  & 0.00 & 36.27 / \textbf{30.23}\\
        \midrule
        \multicolumn{9}{l}{\textbf{Mistral-7B-Instruct-v0.3}}\\
        +No Thought              & 37.18 & 43.71 & 6.25  & 5.58 & 81.52 & 0.00 & 0.00 & 0.00 & 26.59 / 21.78\\
        +CoT                    & 37.18 & 43.05 & 12.50 & 6.59 & 63.04 & 2.04 & 0.00 & 0.00 & 24.85 / 20.55\\
        +Template               & 43.59 & 40.40 & 14.58 & 8.12 & 73.91 & 0.00 & 4.76 & 0.00 & \textbf{27.02 / 23.17}\\
        \midrule
         \multicolumn{9}{l}{\textbf{Mistral-Nemo-12B-Instruct}}\\
        +No Thought            & 34.62 & 66.89 & 10.42 & 7.61 & 82.61 & 0.00 & 9.52 & 0.00 & 32.66 / 26.46\\
        +CoT                    & 65.38 & 60.93 & 20.83 & 7.61 & 83.70 & 0.00 & 9.52 & 0.00 & 35.69 / 31.00\\
        +Template               & 64.10 & 64.90 & 20.83 & 8.63 & 86.96 & 0.00 & 14.29 & 0.00 & \textbf{37.28 / 32.46}\\
        \midrule
         \multicolumn{9}{l}{\textbf{Qwen-2.5-14B-Instruct}}\\
        +No Thought            & 71.79 & 72.84 & 25.00 & 10.15 & 86.92 & 0.00 & 19.05 & 0.00 & 40.74 / 35.72\\
        +CoT                    & 82.05 & 74.83 & 22.92 & 8.63 & 79.35 & 0.00 & 19.05 & 0.00 & 40.75 / 35.85\\
        +Template               & 75.64 & 76.82 & 27.08 & 10.15 & 85.87 & 0.00 & 19.05 & 1.79 & \textbf{42.20 / 37.05}\\
        \bottomrule
    \end{tabular}
    \end{adjustbox}
    \label{tab:nexus_breakdown}
\end{table*}
We report the full Nexus results from our experiments in \autoref{tab:nexus_breakdown}, covering both prompting and fine-tuning strategies across eight dataset sub-categories\footnote{\url{https://huggingface.co/spaces/Nexusflow/Nexus_Function_Calling_Leaderboard}}. Sub-category analysis reveals interesting patterns: while our template-based prompting consistently outperforms baselines in many categories and achieves the best overall performance, we also observe cases where other approaches—such as direct (No-Thought) generation and naive CoT prompting—prove beneficial. We hypothesize that this may be due to prior model exposure to specific functions or APIs, varying task complexity, user query, or differences in documentation structure. These findings further underscore the promise of \textbf{adaptive template} strategies and point to the need for more comprehensive and fair evaluation protocols, which we leave for future work.

\subsection{Prompts For Data Construction}
\label{appendix:data_construction}
We provide prompts for each stage of the \textbf{ToolGT} construction process, including: reasoning chain generation~(\autoref{tab:round_1}), function-call generation using the generated reasoning~(\autoref{tab:round_2}), and training sample filtering~(\autoref{tab:round_3}). Finally, we combine these steps to produce a training example used for fine-tuning, shown in~\autoref{tab:training}.

\begin{table*}[htbp]
\small
\centering
\renewcommand{\arraystretch}{0.3}
\begin{tabular}{p{1cm}|p{12.5cm}}
\toprule
\multicolumn{2}{@{}l}{\textbf{Stage-1: Reasoning
Chains Construction}}\\
\midrule
\textbf{Role} & \textbf{Content} \\ \midrule
\textbf{System} &
You are an expert in composing functions. You are given a question, a set of possible functions and the ground truth function call(s). Based on the question and the ground truth function call(s), you will need to generate the analysis and thought following the given curriculum steps by steps, however, you must pretend that you do not know the ground truth information and assumptions.
If none of the function can be used, point it out. If the given question lacks the parameters required by the function, also point it out.

Here is a list of functions in JSON format that you can invoke.

\textcolor{brown}{\texttt{\{FUNCTIONS HERE\}}}

When composing your analysis, you SHOULD follow this curriculum to have a correct function calling and put your analysis followed this curriculum in <THINKING></THINKING> tags.

\textcolor{brown}{\texttt{\{GUIDED-TEMPLATE HERE\}}}

The output format of all user requests are: 
<THINKING>[Put your thought based on the curriculum step by step here]</THINKING>

\\ \midrule

\textbf{User} &
User request: \textcolor{orange}{\texttt{\{user request\}}}

Ground truth function calling(s):\textcolor{orange}{\texttt{\{GROUND TRUTH\}}}
\\ \midrule
\bottomrule
\end{tabular}
\caption{\textbf{Stage-1 Prompt} during our data construction process, which provides ground truth, template and user query and asking models to generate analysis based on provided information. Note that in this step, we do not allow the models to explicitly say that they know the answer during reasoning generation.}
\label{tab:round_1}
\end{table*}

\begin{table*}[htbp]
\small
\centering
\renewcommand{\arraystretch}{0.3}
\begin{tabular}{p{1cm}|p{12.5cm}}
\toprule
\multicolumn{2}{@{}l}{\textbf{Stage-2: Function-call Generation Using the Generated Reasoning Prompt}}\\
\midrule
\textbf{Role} & \textbf{Content} \\ \midrule

\textbf{System} &
You are an expert in composing functions. You are given a question, a set of possible functions and an analysis to come up with correct function calling(s). Based on the question and provided thinking process, you will need to make one or more function/tool calls to achieve the purpose.
You should only return the function call in tools call sections.
If you decide to invoke any of the function(s), you MUST put it in the format of Put it in the format of [func1(params\_name=params\_value, params\_name2=params\_value2...), func2(params)] in <FUNCTION></FUNCTION> tags

Here is a list of functions in JSON format that you can invoke. 

\textcolor{brown}{\texttt{\{FUNCTIONS HERE\}}}

The output format of all user requests are:

<FUNCTION>[Put function callings (if any) here]</FUNCTION>

\\ \midrule

\textbf{User} &
User request: \textcolor{orange}{\texttt{\{user request\}}}

Given Thinking Process or Analysis:\textcolor{orange}{\texttt{\{ROUND 1 THINKING\}}}
\\ \midrule
\bottomrule
\end{tabular}
\caption{\textbf{Stage-2 Prompt} during our data construction process, which feeds generated reasoning following templates from Stage-1 back into the models to generate final function calling(s). Note that in this step, we do not feed the template into the prompt.}
\label{tab:round_2}
\end{table*}

\begin{table*}[htbp]
\small
\centering
\renewcommand{\arraystretch}{0.3}
\begin{tabular}{p{1cm}|p{12.5cm}}
\toprule
\multicolumn{2}{@{}l}{\textbf{Stage-3: Training Sample Filtering Prompt}}\\
\midrule
\textbf{Role} & \textbf{Content} \\ \midrule

\textbf{System} &
You are an expert in composing functions. You are given a question, a set of possible functions and the ground truth function call(s), and a candidate for the correct function call. Based on the question and the ground truth function call(s), you will need to identify if the provided candidate is also a correct alternative function calling to the ground truth function callings to the question and provided functions. You must pay attention to the functions and parameters were used, if the candidate provide different numbers of functions from the ground truth, it is totally incorrect

Here is a list of functions in JSON format that you can invoke.

\textcolor{brown}{\texttt{\{FUNCTIONS HERE\}}}

You must only output one of the two label

[`CAN REPLACE GROUND TRUTH',`TOTALLY INCORRECT’]. 

The output format of all user requests are: [PUT YOUR LABEL ONLY HERE]

\\ \midrule

\textbf{User} &
User request: \textcolor{orange}{\texttt{\{user request\}}}

Ground Truth Function Calling:\textcolor{orange}{\texttt{\{GROUND TRUTH\}}}

Candidate Function Calling: \textcolor{orange}{\texttt{\{MODEL FUNCTION CALL FROM ROUND 2\}}}
\\ \midrule
\bottomrule
\end{tabular}
\caption{Verification Prompt for Stage-3 during data construction process, which compares ground truth function calling and candidate function callings generated from Stage-2 for alternative solutions apart from ground-truth.}
\label{tab:round_3}
\end{table*}

\begin{table*}[htbp]
\small
\centering
\renewcommand{\arraystretch}{1.3}
\begin{tabular}{p{1cm}|p{12cm}}
\toprule
\multicolumn{2}{@{}l}{\textbf{ToolGT Training Sample}}\\
\midrule
\textbf{Role} & \textbf{Content} \\ \midrule

\textbf{System} &
You are an expert in composing functions. You are given a question and a set of possible functions.

Based on the question, you will need to make one or more function/tool calls to achieve the purpose.

If none of the function can be used, point it out. If the given question lacks the parameters required by the function, also point it out. You should only return the function call in tools call sections.

Here is a list of functions in JSON format that you can invoke 

\textcolor{brown}{\texttt{\{FUNCTIONS HERE\}}}

Should you decide to return the function call(s). 

Put it in the format of [func1(params\_name=params\_value, params\_name2=params\_value2...), func2(params)]

NO other text MUST be included.
\\ \midrule

\textbf{User} &
I'm considering investing and I'd like to know what's happening in the market right now. Could you get me the top market trends in the US?
\\ \midrule
\textbf{Assistant} &
\textbf{<THINKING>}

1. The user is looking for information on current market trends, specifically the top market trends in the US. Based on the provided functions, the "Market Trends API" seems to be the most relevant as it is specifically designed to get the latest market trends.

2. I will select the "Market Trends API" function to fulfill the user's request for market trends in the US.

3. Now, I will examine the documentation for the "Market Trends API" function. The function requires the \texttt{trend\_type} parameter, which specifies the type of trend to retrieve. The parameters \texttt{country} and \texttt{language} are optional but have default values.

4. According to the documentation, the required parameters are:

- \texttt{trend\_type}: This is required and needs to be specified.

- \texttt{country}: This is optional and defaults to "us".

- \texttt{language}: This is also optional, with a default of "en".

5. The user's request does not specify what type of market trend they want, but they mentioned wanting the "top market trends". This implies that they are likely interested in a specific category of trends. Since they did not specify, I will assume a common type of trend like "MARKET\_INDEXES" as a default choice based on the context of "top market trends".

6. The function call that can fulfill the user's request would look like this:

\texttt{Market Trends API(trend\_type="MARKET\_INDEXES", country="us")}

7. I will revalidate the composed function call to ensure it meets the requirements. The \texttt{trend\_type} is specified, and the country defaults to "us" as intended.

8. There are no errors identified in the function call. It satisfies the function documentation and fulfills the user's request for market trends in the US.

\textbf{</THINKING>}

\textbf{<FUNCTION>}

\texttt{[Market Trends API(trend\_type="MARKET\_INDEXES", country="us")]}

\textbf{</FUNCTION>}
\\
\bottomrule
\end{tabular}
\caption{Training sample with reasoning following template used in our \textbf{ToolGT}}
\label{tab:training}
\end{table*}